\documentclass[a4paper,conference]{IEEEtran}
\usepackage{cite}

\ifCLASSINFOpdf
  \usepackage[pdftex]{graphicx}
\else
\fi
\usepackage{amssymb}
\usepackage{amsmath}
\ifCLASSOPTIONcompsoc
\usepackage[caption=false,font=normalsize,labelfont=sf,textfont=sf]{subfig}
\else
  \usepackage[caption=false,font=footnotesize]{subfig}
\fi
\usepackage{url}

\usepackage{makecell}
\usepackage{flushend}
\begin{document}
%
\title{Building Computationally Efficient and Well-Generalizing Person Re-Identification Models with Metric Learning}

\author{\IEEEauthorblockN{Vladislav Sovrasov}
\IEEEauthorblockA{Lobachevsky State University of Nizhni Novgorod, Russia\\
IOTG Computer Vision (ICV), Intel\\
Email: sovrasov.vladislav@itmm.unn.ru}
\and
\IEEEauthorblockN{Dmitry Sidnev}
\IEEEauthorblockA{IOTG Computer Vision (ICV), Intel\\
Email: dmitry.sidnev@intel.com}
}


%


\maketitle

\begin{abstract}
  This work considers the problem of domain shift in person re-identification.
  Being trained on one dataset, a re-identification model usually performs much
  worse on unseen data. Partially this gap is caused by the relatively small
  scale of person re-identification datasets (compared to face recognition ones,
  for instance), but it is also related to training objectives. We propose to
  use the metric learning objective, namely AM-Softmax loss, and some additional
  training practices to build well-generalizing, yet, computationally efficient
  models. We use recently proposed Omni-Scale Network (OSNet) architecture
  combined with several training tricks and architecture adjustments to obtain
  state-of-the art results in cross-domain generalization problem on a
  large-scale MSMT17 dataset in three setups: MSMT17-all$\rightarrow$DukeMTMC,
  MSMT17-train$\rightarrow$Market1501 and MSMT17-all$\rightarrow$Market1501. Training code
  and the models are available online in the GitHub repository\footnote{
    \url{https://github.com/opencv/openvino_training_extensions/tree/develop/pytorch_toolkit/object_reidentification/person_reidentification}}.

\end{abstract}


%
\IEEEpeerreviewmaketitle

\section{Introduction}
Recently, deep learning approaches have taken leading positions in many
computer vision tasks such as image classification, object detection, semantic
segmentation, face recognition, person re-identification
\cite{alexnet,Girshick2013RichFH,Shelhamer2014FullyCN,GoogleFaceNet,firstCNNReid}
and many others.  Modern ways to solve these tasks are based on convolutional
neural networks (CNN).  Common weak point of CNNs is domain shift
\cite{baselineCrossDomain,segmDomainShift}. The impact of this problem depends
on the nature of the computer vision task and diversity of available training
data.  The person re-identification task is to build a discriminative
identity-preserving person descriptor for performing large-scale person retrieval
from diverse video streams coming from different cameras under varying lighting
and background conditions. Latest advances in face recognition
\cite{deng2018arcface,amSoftmax} can make one think that domain shift is not a
significant issue for person re-identification as well, since both tasks are
solved using similar approaches of building identity-preserving
descriptors. Unfortunately, performance of person re-identification CNNs
drastically degrades on unseen data captured under different (compared to the
training data) conditions. Many works are devoted to the cross-domain adaptation
problem \cite{baselineCrossDomain,inDefenceTriplet2,yuan2019calibrated} because
this question is critical for practical use of re-identification models.

We believe that huge domain shift problem experienced by the current
state-of-the-art person re-identification models is due to the following
reasons:

\begin{itemize}
  \item Relatively small size of the available datasets (see statistics in Table
    \ref{tab:data}). Currently the largest and the most challenging public
    person re-identification dataset is MSMT17 \cite{MSMT17}. It consists of
    $126142$ images and contains $4101$ identities. At the same time, the most
    popular face recognition dataset MS-Celeb-1M \cite{ms1m} has about $10$M
    images of $1$M identities. Such a huge scale allows models trained on
    the MS-Celeb-1M to be robust in benchmarks containing unseen data
    \cite{deng2018arcface} from other domains.
  \item The nature of the person re-identification problem itself: appearance of
    a person seems to be not as discriminative as appearance of their face. For
    example, distinguishing two people in similar dark clothes from the back side
    can be challenging even for a human.
  \item A common way of building re-identification models is to use ResNet-50
    \cite{resnet} as a backbone with the cross-entropy loss function
    \cite{baselineCrossDomain, hpm}. Given the small amount of the training data
    it tends to overfit and provides poor discriminative qualities. To some
    extent it is alleviated by regularization techniques and/or various loss
    functions \cite{mgn, abd, lmSoftmax, centerLoss}, but there are more
    effective ways to do that.
\end{itemize}

\begin{table}
  \caption{Statistics of Large-scale Person Re-identification Datasets}
  \label{tab:data}
  \centering
  \begin{tabular}{l|c|c|c}
    Dataset & IDs & Images & Cameras \\ \hline
    VIPeR \cite{Gray2007EvaluatingAM} & 632 & 1264 & 2 \\
    GRID \cite{Loy2009MulticameraAC} & 251 & 1275 & 6 \\
    CUHK01 \cite{cuhk01} & 971 & 3882 & 2 \\
    CUHK03 \cite{firstCNNReid} & 1467 & 28192 & 2 \\
    \hline
    Market1501 \cite{market} & 1501 & 32668 & 6 \\
    DukeMTMC-ReID \cite{duke, duke2} & 1812 & 36411 & 8 \\
    MSMT17 \cite{MSMT17} & 4101 & 126411 & 15 \\
    \hline
  \end{tabular}
\end{table}

In this work we use AM-Softmax \cite{amSoftmax} loss within OSNet
\cite{zhou2019osnet} architecture to get discriminative features that allow us
to obtain fast cross-domain networks that generalize well to unseen data.

In brief, the key contributions of this paper can be summarized as follows:

\begin{itemize}
  \item We apply the metric-learning approach (AM-Softmax \cite{amSoftmax} loss)
    to further improve generalization of re-identification models in the
    cross-domain setup.

    \item We explore some additional training tricks and modifications of the
    OSNet to make lightweight models that are strong enough in the cross-domain
    setup and require significantly less computations than the top-performing
    ones.

  \item Our combination of adjustments to OSNet obtains SOTA results in three
    cross domain setups MSMT17-all$\rightarrow$DukeMTMC,
    MSMT17-train$\rightarrow$Market1501, MSMT17-all$\rightarrow$Market1501
    among approaches that do not use data from
    the target domains in training.

  \item In order to see which models are suitable for real-time applications, we
    evaluate the performance of the developed models on a desktop CPU.

\end{itemize}

\begin{figure*}[ht]
  \centering
  \subfloat[Activation maps of OSNet-IAP 1.0x trained with the Softmax loss]
  {{\includegraphics[width=.65\textwidth]{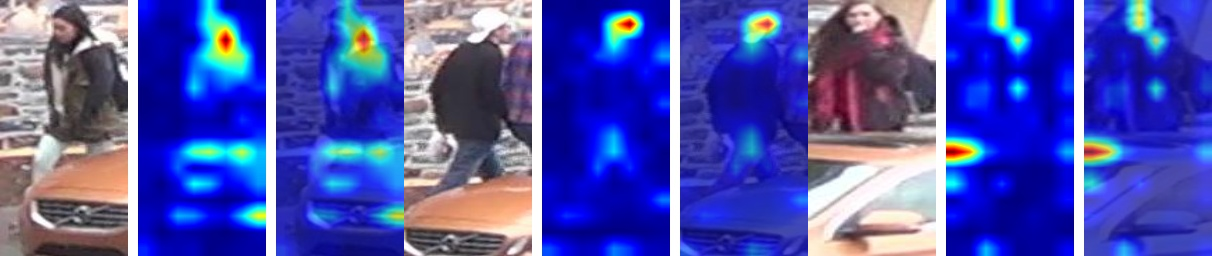}}}

  \subfloat[Activation maps of OSNet-IAP 1.0x trained with the AM-Softmax loss]
  {{\includegraphics[width=.65\textwidth]{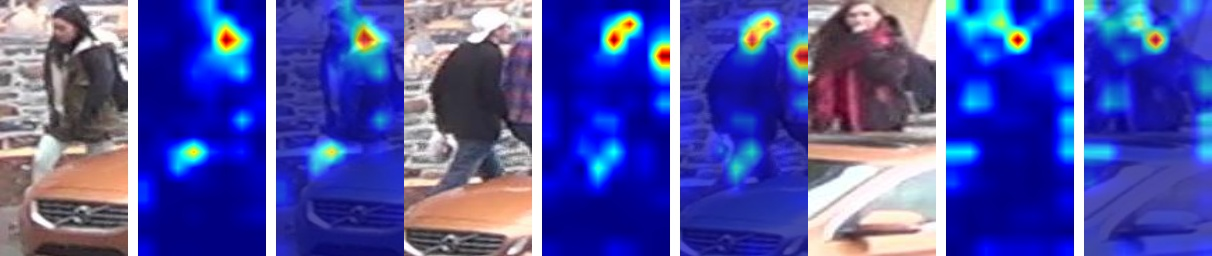}}}
  \caption{The difference between activations at the last feature map of the models trained with
           Softmax and AM-Softmax. The model with Softmax pays attention to the car in all the presented cases
           while AM-Softmax-based model fails to discriminate car and person only in one case (on the third map from the left).
           All the images are taken from DukeMTMC-ReID, models are trained on MSMS17-all}
  \label{fig:actmaps}
\end{figure*}

\begin{figure*}
  \centering
  \subfloat[Top10 images retrieved by OSNet-IAP 1.0x trained with the Softmax loss]
  {{\includegraphics[width=.65\textwidth]{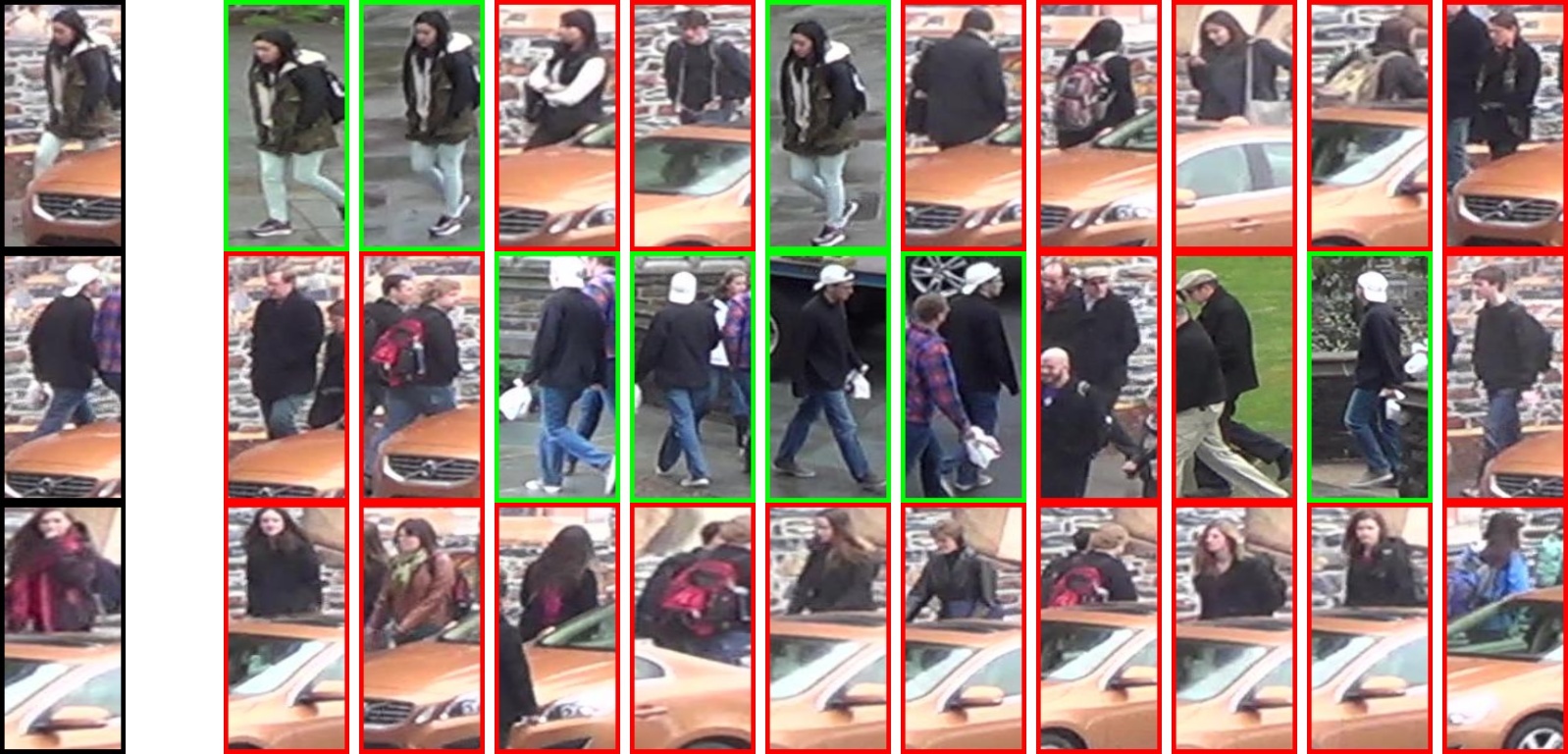}}}

  \subfloat[Top10 images retrieved by OSNet-IAP 1.0x trained with the AM-Softmax loss]
  {{\includegraphics[width=.65\textwidth]{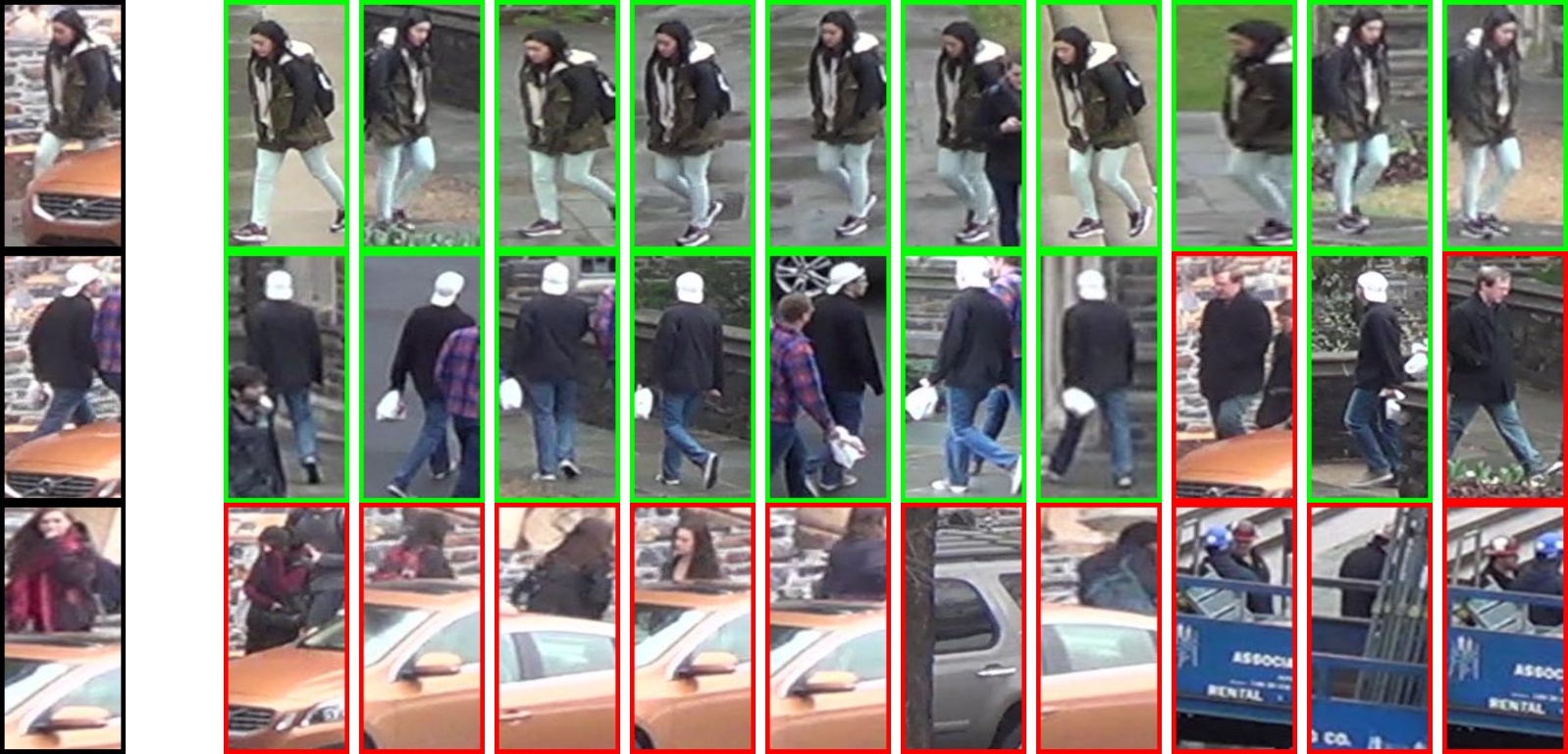}}}
  \caption{Retrieving results from the models trained with Softmax and AM-Softmax.
    Query images are on the left side. Incorrect matches are enclosed by red rectangles.
    The model trained with Softmax generates corrupted activations on all three
    query images (see Figure \ref{fig:actmaps}), that gives a lot of false
    matches containing the orange car. AM-Softmax-based model gives near-perfect
    Top-10 in the first two cases, but fails in the third one.  All the images
    are taken from DukeMTMC-ReID, the models are trained on MSMS17-all}
  \label{fig:top10}
\end{figure*}

\section{Related Work}

\paragraph{Efficient CNN architectures built for person re-identification}

There is a wide variety of approaches to person re-identification, but to our
best knowledge, only a few of them aim to develop a CNN architecture specifically designed for
the re-id task from scratch \cite{izutov2018fast, zhou2019osnet}.  RMNet
proposed in \cite{izutov2018fast} is a lightweight and computationally efficient
architecture aimed to work on low-power devices. It is based on the ResNet
\cite{resnet} paradigm mixed with other techniques for building efficient CNNs
\cite{mobilenet, squeezenet}.  This network trained by a carefully designed
procedure reaches similar results to the relatively heavy ResNet-50-based
alternatives like MGN \cite{mgn} and HPM \cite{hpm}.

The next generation of efficient architectures for person re-identification is
based on Res2Net \cite{res2net} multi-scale paradigm and presented in
\cite{zhou2019osnet}. It also employs insights from the lightweight
architectures \cite{mobilenet, mobilenetv2} and SE blocks \cite{seblock} to
build efficient multi-scale residual blocks.

Later, OSNet has been adapted for the cross-domain
scenario \cite{zhou2019learning} by incorporating Instance normalization \cite{instance} (IN)
layers. The optimal placement of INs is found by the NAS
\cite{Dong2019SearchingFA} technique. The resulting OSNet-AIN shows great
cross-domain generalization outperforming some of the recently proposed
label-free methods that use target data \cite{Yu2019UnsupervisedPR,
  Yang2019PatchBasedDF}.

\paragraph{Loss functions for person re-identification}

The person re-identification task aims to find a mapping function that
translates the image domain into the high-dimensional vector space
$f:I\to\mathbb{R}^N$ such that images of the same person captured under
different conditions become close in the vector space. At the same time, vectors
corresponding to different persons should be far from each other.  The most
popular way to address this problem is to treat it as multi-class classification
and solve using the cross-entropy (CE) loss
\cite{zheng2016person,Xiao2016LearningDF}.  Later, it has been explored that CE
favors separability of features rather than discriminative properties
\cite{lmSoftmax, centerLoss}. To overcome this drawback,
several margin-based variations of the CE loss were introduced
\cite{amSoftmax,deng2018arcface}.  Despite advantages of the margin-based
softmax losses, they are not widespread yet in the person re-identification
field and only several works use them for single domain training \cite{SphereReID,
  izutov2018fast, Luo2018SpectralFT}.

Alternative approach to learn a discriminative mapping is to use the triplet
loss \cite{inDefenceTriplet1, inDefenceTriplet2}.  This approach requires
carefully designed hard mining strategies due to a vast amount of possible
triplets. Often the triplet loss is added as a complementary one to cross entropy
\cite{mgn,Luo2019BagOT,abd}.

\paragraph{Cross domain re-identification}

After emergence of several large-scale person re-identification benchmarks
\cite{market, duke2, MSMT17} the problem of domain shift between them has become
clearly visible.  Domain adaptation, unsupervised and cross-domain methods have
been designed to handle this problem.

Domain adaptation approach assumes availability of labels on source domain and
unlabeled data from the target one. The training is performed on a mix of
labeled and unlabeled data. And the evaluation is performed on the target domain
(if test data from the target domain exists).  Former domain adaptation methods
utilized GANs to generate new data with the distribution similar to the target
domain in offline or online mode
\cite{Zhong2018GeneralizingAP,deng2017imageimage,MSMT17}.  Latter approaches try
to adapt to source domain via solving auxiliary tasks on the target data
\cite{Huang2018EANetEA} or to use unsupervised methods
\cite{Fu2018SelfsimilarityGA, yu2017crossview, Fan2017UnsupervisedPR} that don't
require labels at all.

Although fully unsupervised methods are attractive, there is a performance gap
between them and supervised approaches \cite{ye2020deep}.  So, from the
practical perspective, only supervised cross-domain methods are suitable for
real life applications at this moment. Besides, once trained, well-generalizing
model can be deployed without any retraining.

Cross-domain generalization is addressed by adjusting the architecture of CNNs
\cite{zhou2019learning, Jia2019FrustratinglyEP}, designing task-specific loss
functions \cite{inDefenceTriplet2} or adversarial training
\cite{yuan2019calibrated}.  Improvements in each of this fields can be
transferred to unsupervised or semi-supervised methods since many of them use
parts of supervised training procedures under the hood.  Our work pays attention
to the loss function, model architecture and details of the training process
(augmentation, schedules, data sampling) to build better cross domain models.

\section{Method}

\subsection{Loss function for deep metric learning}

To learn the mentioned identity-preserving mapping function $f:I\to\mathbb{R}^N$
there are currently two main approaches: to use an identity classification loss
\cite{zheng2016person, Xiao2016LearningDF} acting as a global rule or use local rules such as triplet
loss \cite{inDefenceTriplet1, inDefenceTriplet2}. Both approaches are
sub-optimal and currently the best results are obtained by its combination
\cite{mgn, abd, Luo2019BagOT}. The purpose of the triplet loss in the mentioned
papers is to fix a poor ability of the CE to learn a well-generalizing mapping.
CE loss can perform well in the single-domain setup, because training and
testing subsets of public datasets \cite{market, duke2, MSMT17} are very similar
(they contain randomly sampled identities from the same scenes).  But in cross
domain setup we should have a stronger and more structured supervision to
control the properties of $f$. The family of angular margin-based losses
(SphereFace\cite{Liu2017SphereFaceDH}, AM-Softmax\cite{amSoftmax},
ArcFace\cite{deng2018arcface}) allows us to achieve that. We use AM-Softmax
since it provides the required properties and is easier to
optimize than SphereFace or ArcFace even on noisy data.  AM-Softmax is defined
by the following formula:

\begin{gather}
    L_{ASM} = -\sum_i \log p_i, \\
    p_i=\frac{e^{s(W^T_{y_i}f_i-m)} }{e^{s(W^T_{y_i}f_i-m)} + \sum_{j,j\ne y_i}e^{sW^T_{y_j}f_j} },
    \label{eq:ams}
\end{gather}

where $f_i\in\mathbb{R}^N$ are $l_2$-normalized outputs of the mapping $f$,
$W\in\mathbb{R}^{N\times M}$ are $l_2$-normalized weights of the linear layer
transforming $f_i$ to the space of logits, $m$ and $s$ are a margin between
classes and a scale of features.  Non-zero margin forces the loss not only to
make vectors $f_i$ closer to their prototypes $W^T_{y_i}$ in terms of the cosine
distance, but also to create margins between different classes. That process
makes the mapping $f$ discriminative. The scale parameter $s$ controls the degree of similarity
between $f_i$ and $W^T_{y_i}$ required to generate sharp distribution $p_i$.
High value of $s$ corresponds to the case, when the similarity between $f_i$ and $W^T_{y_i}$
should be just slightly greater, than between $f_i$ and $W^T_j,j\ne y_i$ to get close to one-hot distribution $p_i$.

Since AM-Softmax loss defines a strong global placement rule for vectors $f$ on
a high-dimensional hypersphere, it needs to be slightly relaxed to prevent
overfitting. Authors of \cite{Adaimi2019RethinkingPR} evaluate several
approaches to decrease the sharpness of the distribution $p_i$ (\ref{eq:ams}) on easy samples
and simple subtraction of entropy from the softmax loss gives them best
results. Taking that into account, we will define the AM-Softmax-based identity
loss as:

\begin{equation}
  \label{eq:id_loss}
  L_{id} = [L_{ASM} + \alpha \sum_i p_i \log p_i]_+,
\end{equation}
where $p_i$ are from (\ref{eq:ams}).

\subsection{Model architecture}

As stated earlier, we use OSNet as a baseline CNN architecture for the person
re-identification task. Like Res2Net, it provides a multi-scale residual block,
but at the same time, OSBlock is lightweight.  Compared to the standard residual
block, OSBlock has larger theoretical receptive field. Hence, it can provide
aggregation and processing of global context starting from shallow layers. It
seems to be the major contributing factor that allows OSNet to outperform many
ResNet-50 based approaches \cite{zhou2019osnet} without any sophisticated
training tricks, requiring about 2.0 billions of the floating point operations
while ResNet-50 requires 5.3 billions for the same input resolution
$256\times128$.

OSNet already has a well-balanced architecture, so there are only minor
adjustments from our side:

\begin{itemize}
  \item By default, OSNet uses a global average pooling operation to aggregate
    spatial features into a vector. We replace it with the global depthwise
    convolution \cite{mobilefacenet}.  It allows us to make aggregation of the
    final feature map more flexible since each channel and each spatial position
    has a learnable weight instead of uniform weights in case of the average
    pooling.  Global depthwise convolution also slightly increases the capacity
    of the lightweight models without introducing noticeable overhead.

  \item It has been explored that InstanceNorms can boost the performance in
    cross-domain re-identification \cite{Jia2019FrustratinglyEP}. Following this
    practice, we insert InstanceNorms before the first convolution and after it
    (instead of BatchNorms) at least to decrease the color distribution shift.

\end{itemize}

AM-Softmax loss treats the normalized representation vectors $f_i$ as points on
the hypersphere $||f||_{l_2}=1$. Such points can have both positive and
negative-signed coordinates.  To let the model produce the output vector with
negative components too, we add PReLU activation layer instead of ReLU in the
original OSNet. Also we use 256-dimensional output layer, while the original
OSNet comes with a 512-dimensional one. AM-Softmax generates more structured
representation than Softmax, so we can learn a compact embedding space. As a
bonus, low-dimensional embeddings lead to faster distance computation, averaging
and any other operations performed on the extracted embeddings.

We refer to the modified OSNet architecture as OSNet-IAP.

\section{Experiments}

\begin{table*}
\caption{Cross-domain Results Obtained by OSNet-IAP in Different Setups}
\label{tab:cross}
  \centering
  \begin{tabular}{l|c|c|c|c|c|c}
    \hline
    Model, train data & \multicolumn{2}{c|}{MSMT17-test} & \multicolumn{2}{c}{Market1501} & \multicolumn{2}{|c}{DukeMTMC-ReID}   \\
    & \textit{Rank-1} & \textit{mAP} & \textit{Rank-1} & \textit{mAP}  & \textit{Rank-1} & \textit{mAP}  \\
    \hline
    OSNet-IAP 1.0x, MSMT17-train & 77.97 & 48.66 & 69.27 & 40.27 & 64.00 & 42.16 \\
    \hline
    OSNet-IAP 0.25x, MSMT17-all & - & - & 72.09 & 42.14 & 62.97 & 41.86 \\
    OSNet-IAP 0.5x, MSMT17-all & - & - & 77.49 & 48.98 & 70.92 & 50.87 \\
    OSNet-IAP 0.75x, MSMT17-all & - & - & 79.78 & 52.49 & 73.83 & 53.85 \\
    OSNet-IAP 1.0x, MSMT17-all & - & - & 82.66 & 55.70 & 74.91 & 56.6 \\
    \hline
    OSNet-IAP 1.0x, MSMT17-all + DukeMTMC-ReID-all & - & - & 83.52 & 58.02 & - & - \\
    OSNet-IAP 1.0x, MSMT17-all + Market1501-all & - & - & - & - & 76.66 & 58.93\\
    \hline
    OSNet-IAP 1.0x, MSMT17-all + Private data & - & - & 85.87 & 58.58 & 77.06 & 58.59 \\
    \hline
  \end{tabular}
\end{table*}

\begin{table*}
\caption{Comparison with the Current State-of-the-art Methods in the Cross-domain Re-identification}
\label{tab:sota}
  \begin{minipage}[c]{\hsize}
  \centering
  \begin{tabular}{l|l|cc|cc}
    \hline
    Method & Train data & \multicolumn{2}{c}{Market1501} & \multicolumn{2}{|c}{DukeMTMC-ReID}   \\
     & & \textit{Rank-1} & \textit{mAP}  & \textit{Rank-1} & \textit{mAP}  \\
    \hline
    ABD-Net \cite{abd} (single domain) \footnote{Measured by us using the official
    model and evaluation script from \url{https://github.com/TAMU-VITA/ABD-Net}}
    & MSMT17-train & 50.15 & 25.95 & 48.7 & 29.9 \\
    \hline
    CE-FAT \cite{inDefenceTriplet2} & MSMT17-train & 52.80 & 25.40 & 50.90 & 31.30 \\
    ADIN \cite{yuan2019calibrated} & MSMT17-train & 59.10 & 30.30 & 60.70 & 39.10 \\
    CDB \cite{baselineCrossDomain} & MSMT17-train & 64.80 & 36.60 & 64.50 & 43.3 \\
    ADFL \cite{Liu2019AttentionAB} & MSMT17-train & 68.00 & 37.70 & \textbf{66.30} & \textbf{46.20} \\
    \hline
    OSNet-IAP 1.0x (Ours) & MSMT17-train & \textbf{69.27} & \textbf{40.27} & 64.00 & 42.16 \\
    \hline
    \hline
    OSNetx1.0-IBN \cite{zhou2019osnet} & MSMT17-all & 66.50 & 37.20 & 67.40 & 45.60 \\
    MAR \cite{Yu2019UnsupervisedPR} & MSMT17-all+Market(U)/Duke(U) & 67.70 & 40.00 & 67.10 & 48.00 \\
    OSNetx1.0-AIN \cite{zhou2019learning} & MSMT17-all & 71.10 & 52.70 & 70.10 & 43.30 \\
    PAUL \cite{Yang2019PatchBasedDF} & MSMT17-all+Market(U)/Duke(U) & 68.50 & 40.10 & 72.00 & 53.20 \\
    \hline
    OSNet-IAP 0.75x (Ours) & MSMT17-all & 79.78 & 52.49 & 73.83 & 53.85 \\
    OSNet-IAP 1.0x (Ours) & MSMT17-all & \textbf{82.66} & \textbf{55.70} & \textbf{74.91} & \textbf{56.60} \\
    \hline
  \end{tabular}
  \end{minipage}
\end{table*}

\begin{figure}[ht]
  \begin{center}
    \includegraphics[width=0.8\linewidth]{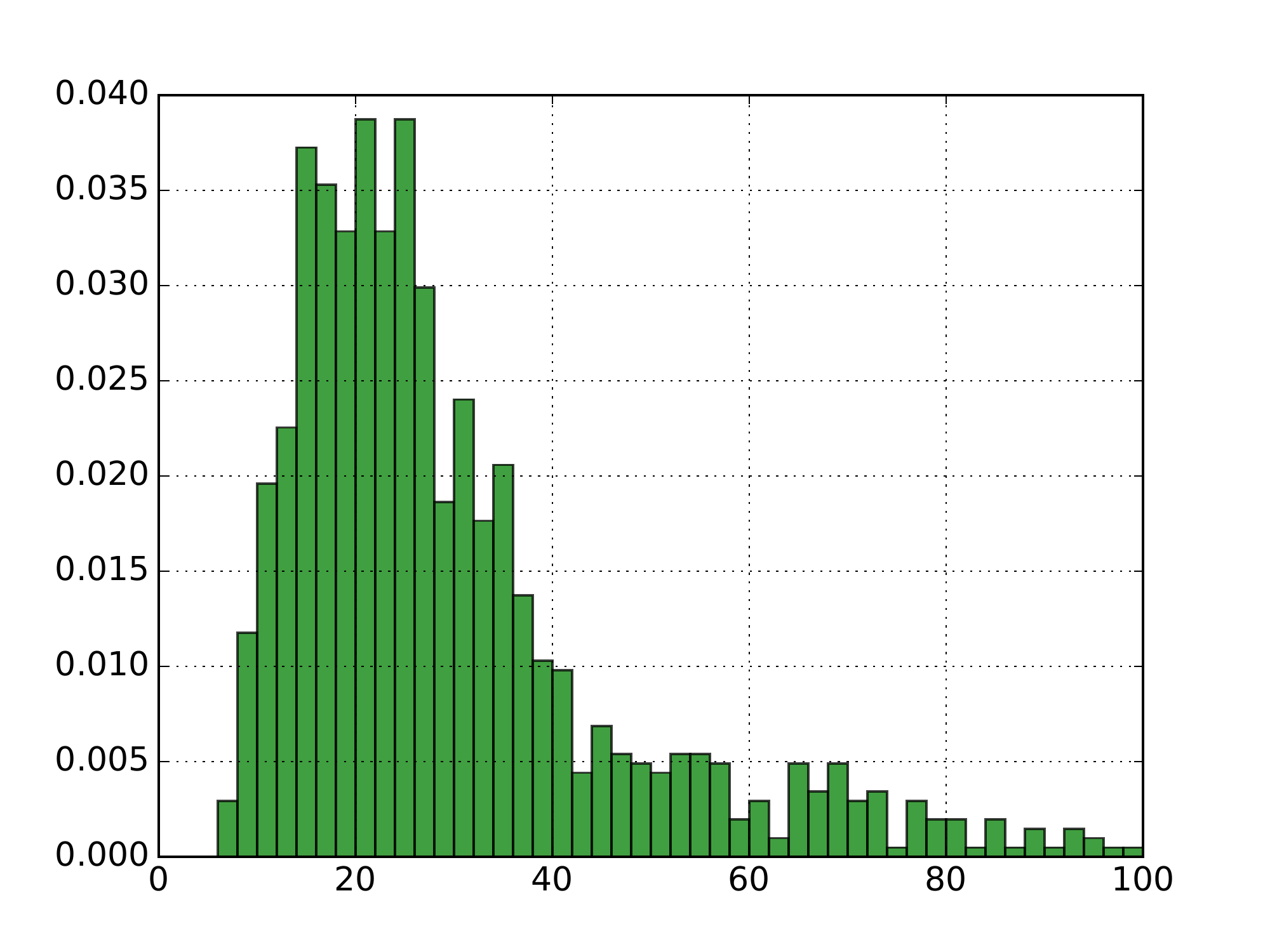}
  \end{center}
  \caption{Distribution of the amount of images per identity in MSMT17 dataset}
  \label{fig:msmt_hist}
\end{figure}

\begin{table}
\caption{Same Domain Performance of Cross-domain Methods on MSMT17}
\label{tab:same}
  \centering
  \begin{tabular}{l|cc}
    \hline
    Method & \textit{Rank-1} & \textit{mAP} \\
    \hline
    ABD-Net \cite{abd} (single domain) & 82.30 & 60.80 \\
    \hline
    CE-FAT \cite{inDefenceTriplet2} & 69.40 & 39.20 \\
    ADFL \cite{Liu2019AttentionAB} & \textbf{78.20} & \textbf{48.80} \\
    \hline
    OSNet-IAP 1.0x (Ours) & 77.97 & 48.66 \\
    \hline
  \end{tabular}
\end{table}

\begin{table*}
  \centering
  \caption{Ablation Study on MSMT17-all Dataset}
  \label{tab:ablation_ms}
  \begin{tabular}{l|cc|cc}
    \hline
    Model & \multicolumn{2}{c}{Market1501} & \multicolumn{2}{|c}{DukeMTMC-ReID}   \\
     & \textit{Rank-1} & \textit{mAP}  & \textit{Rank-1} & \textit{mAP}  \\
    \hline
    OSNet 1x with Softmax & 67.04 & 38.65 & 65.53 & 46.78 \\
    + Random flip \& Color jitter & 71.38 & 37.07 & 67.10 & 46.88 \\
    + AM-Softmax & 74.70 & 46.47 & 69.52 & 49.60 \\
    + Depthwise pooling & 75.18 & 46.55 & 70.38 & 50.16 \\
    + Instance Norms & 75.53 & 46.96 & 70.92 & 50.91 \\
    + Uniform identity sampling & 81.44 & 53.93 & 73.38 & 55.06 \\
    + Advanced augmentations & 82.66 & 55.70 & 74.91 & 56.60 \\
    \hline
  \end{tabular}
\end{table*}

\begin{table}
  \centering
  \caption{Average Pairwise Cosine Distance between Centroids Corresponding to 200 Randomly
           Picked Identities from Different Datasets. Models are Trained on MSMT17-train}
  \label{tab:cosine}
  \begin{tabular}{l|cc}
    \hline
    Dataset & \makecell{OSNet-IAP 1.0x \\ w/o AM-Softmax \\ + Softmax} & OSNet-IAP 1.0x  \\
    \hline
    Market-1501   & 0.35 & 0.71 \\
    DukeMTMC-ReID & 0.36  & 0.81  \\
    \hline
    MSMT17-test   & 0.41 & 0.91   \\
    \hline
  \end{tabular}
\end{table}

\begin{table*}
\caption{Performance on the Intel\textregistered  Core\texttrademark i7-6700K 4.00GHz CPU in OpenVINO\texttrademark R3 2019 Toolkit.
         Batch Size is Set to 1, Input Resolution is $256\times128$, Inference Precision is FP32}
\label{tab:perf}
  \centering
  \begin{tabular}{l|c|c|c|c}
    Model & GFLOPs & Parameters, M & FPS & Latency, ms \\ \hline
    ResNet-50 & 5.30 & 23.50 & 74.51 & 13.42 \\
    OSNet 1.0x & 1.99 & 2.05 & 162.90 & 6.13 \\
    OSNet-IAP 1.0x & 1.99 & 2.12 & 157.64 & 6.34 \\
    OSNet-IAP 0.75x & 1.17 & 1.24 & 250.65 & 3.99 \\
    OSNet-IAP 0.5x & 0.56 & 0.60 & 441.64 & 1.26 \\
    OSNet-IAP 0.25x & 0.17 & 0.18 & 911.93 & 1.09 \\
    \hline
  \end{tabular}
\end{table*}

\paragraph{Datasets and cross-domain training issues}

For training and evaluation we use 3 largest publicly available datasets: MSMT17
\cite{MSMT17}, Market1501 \cite{market} and DukeMTMC-ReID \cite{duke, duke2}.
Their statistics are shown in the Table \ref{tab:data}. Among them, MSMT17 has
drastically more identities and images. We are focused on large scale data, thus
we use MSMT17-train as a core data subset for all of our experiments and add
MSMT17-test, Market1501 or DukeMTMC-ReID depending on setup.

In single domain setup, better performance on training data, generally, means
better performance on the test (given that we avoid overfitting). For the
cross-domain scenario that's not always true. For instance, if the majority
of identities presented in the training split wear shorts, we shouldn't expect
great performance on a test split where most people wear trousers, since we have
a drastic appearance bias in these two datasets. In case of large-scale data
this problem is not so severe, but the model can still capture less obvious
dataset-specific details or noise. In the cross-domain setup we have to make
models slightly underfitted to provide higher performance on unseen data.

\paragraph{Training strategy}

To avoid overfitting we use intensive data augmentations with a long training
schedule.  We train all models with AMSGrad optimizer \cite{amsgrad} for 65
epochs. The initial learning rate is $0.0015$.  It's dropped by a factor of $10$
twice: at the epochs 40 and 50.  Values of the scale $s=0.3$ and margin $m=0.35$ are taken
from the original paper \cite{amSoftmax} without changes. Regularization
parameter $\alpha$ from (\ref{eq:id_loss}) is set to $0.3$, although, the model
seems to be insensitive to this parameter in the range $\alpha\in[0.1,0.5]$.
Batch size is 64. Each batch is sampled from 16 randomly chosen identities, totally
we have 4 images per identity in a batch. If one or several of the sampled identities
have less than 4 images, additional identities are randomly picked to
complete the batch. We found this sampling strategy to be effective in combination with the AM-Softmax
loss. It allows us to partially alleviate imbalance in the amount of images per
identity (see Figure \ref{fig:msmt_hist}).  Initial weights are taken from the
original OSNet pre-trained on the ImageNet \cite{imagenet_cvpr09}.  Also we use
a kind of warm-up proposed in \cite{zhou2019osnet}: for the first 5 epochs the
base network pre-trained on ImageNet is frozen and only the randomly initialized
classifier and depthwise global convolution layers are trained.  Data augmentation
includes random color transformations (jittering in the HSV space, conversion to
grayscale), spatial deformations (random rotation, horizontal flip, padding) and
image distortion (random erase \cite{zhong2020random}, drawing of random figures
and grids). In case of training on MSMT17-train set we observed overfitting.
To handle that we add  Gaussian Continuous Dropout layers \cite{continuosDropout} with
$\mu=0.1,\:\sigma=0.03$ after the convolutional branch of the each OSBlock.
We don't use this regularization when training on larger sets.

Our training code is based on the Torchreid library
\cite{torchreid} and available as a part of the
OpenVINO\textsuperscript{\textregistered} Training Extensions toolkit
\footnote{\url{https://github.com/opencv/openvino_training_extensions/}}.

\subsection{Results in cross-domain person re-id}

We evaluate the scalability of OSNet-IAP in data and model size dimensions (see
Table \ref{tab:cross}).  Following the original OSNet, we use multiplier
$\beta\in\{0.25,0.5,0.75,1.0\}$ to control the amount of channels. OSNet-IAP
reacts smoothly to varying $\beta$. The most dramatic performance drop occurs
between $\beta=0.5$ and $\beta=0.25$.

Evaluation subset of MSMT17 is 3x larger than the training one. Combining train
and test subsets of MSMT17 gives OSNet-IAP additional $~6\%$ to rank-1 on
Market1501 and almost $11\%$ on DukeMTMC-ReID. OSNet-IAP 0.25x performs at the
same level as OSNet-IAP 1.0x trained with 4x fewer amount of the data. Combining
MSMT17-all with DukeMTMC-ReID brings a significant improvement on Market1501 as
well as combining MSMT17-all and Market1501 gives a clear boost on
DukeMTMC-ReID. We've also collected a private person re-identification dataset
containing 216284 images of 7187 identities. The dataset is built from
single-camera scenes and thus it is not so as representative as MSMT17. But,
still, when we combine this data with MSMT17 we observe a clear boost on both
DukeMTMC-ReID and Market1501 domains. This indicates that OSNet-IAP can further
benefit from a more diverse dataset and we believe that increasing the amount
of public carefully designed large-scale datasets will push the limits of cross-domain
generalization in person re-identification task to a new level.

\paragraph{Comparison with the state-of-the-art}\label{para:compare_sota}

Recently, cross-domain methods demonstrated an impressive progress in zero-shot
transfer. ADFL \cite{Liu2019AttentionAB} incorporates attention mechanism into ResNet-50 and
shows great results on Market1501 and DukeMTMC-ReID using only MSMT17-train as the source data. Even still,
OSNet-IAP 1.0x slightly outperforms it in MSMT17-train$\rightarrow$Market1501
setup (see Table \ref{tab:sota}). SOTA-level single domain model
ABD-Net \cite{abd} demonstrates worse performance than any of the considered cross-domain methods,
although using the same training data.

With larger training set, OSNet-IAP 1.0x significantly improves performance in
both MSMT17-all$\rightarrow$Market1501, MSMT17-all$\rightarrow$DukeMTMC-ReID
setups outperforming the original cross-domain variations of OSNet (IBN and
AIN).  At the same time, OSNet-IAP surpasses unsupervised domain adaptation
approaches that use unlabeled target data and MSMT17 with labels as an auxiliary
dataset (for pre-training or assigning soft labels, etc.).

Cross-domain model should perform equally well in all domains including the
target one. SOTA-level cross-domain models demonstrate good performance on the
target data, although their results are significantly lower than the highest
score in the same domain setup (see Table \ref{tab:same}).  OSNet-IAP 1.0x shows
better performance on target (MSMT17) domain than on unseen ones (Market1501 and
DukeMTMC-ReID, see Table \ref{tab:sota}), yielding the first place to
ADFL. Thus, OSNet-IAP outperforms ADFL in MSMT17-train$\rightarrow$Market1501,
but yields in MSMT17-train$\rightarrow$DukeMTMC-ReID and
MSMT17-train$\rightarrow$MSMT17-test setups.  Considering all of the above,
OSNet-IAP tends to be underfitted on MSMT17-train (this leads to worse score on
DukeMTMC-ReID, since it's somewhat similar to MSMT17) and more biased towards
Market1501 than ADFL.

\subsection{Ablation study}
\label{subsec:ablation}

In this section we'll evaluate the contribution of each of the training tricks
and architecture adjustments to the final cross-domain score of OSNet-IAP. We
take OSNet 1x with Softmax without augmentations as a baseline, MSMT17-all as a
source domain, Market1501 and DukeMTMC-ReID as target domains. Results are shown
in Table \ref{tab:ablation_ms}.  Each component slightly increases the overall
metric, but the most noticeable improvement is due to the AM-Softmax loss, basic
augmentations (flip, color jitter) and sampling strategy.  Color augmentation is
a crucial step in the cross-domain training pipeline, since it allows us to mitigate the gap
between color distributions of different domains. Instance Normalization of
input data at shallow layers also helps in that. AM-Softmax-based model
noticeably benefits from the sampling of uniform number of images per identity
in each batch.  This technique allows to handle differences in the number of
samples per identity in the training dataset: all identities have equal
probability to appear in a batch instead of having a correlation with the number
of samples representing an identity.

To prove the actual discriminative ability of AM-Softmax-guided features, we
randomly picked sets of 200 well-represented identities that have 20 or more
images from several datasets.  Then we extracted normalized embeddings using
OSNet-IAP 1.0x trained on MSMT17-train with and without AM-Softmax, computed
centroids for each identity and estimated average pairwise cosine distance
between the obtained centroids. Results of this experiment are presented in the
Table \ref{tab:cosine}. AM-Softmax generates wide distance margins between
centroids, corresponding to different identities, even on
unseen domains (Market1501 and DukeMTMC-ReID). On the contrary, softmax-based model creates
narrow distance margins between centroids, even on the data from the source domain (MSMT17-test).
For all the evaluated models the average inter-centroid distance on the source domain is greater than on
unseen ones. Thus, the average pairwise distance between centroids could be considered
as a similarity measure between domains. Model fitted to the MSMT17 distribution splits different identities
from DukeMTMC-ReID better than ones from Market1501. This fact proves that
MSMT17 is more similar with DukeMTMC-ReID than with Market1501.

Training with AM-Sofrmax also leads to better discrimination between background
and foreground without introducing any explicit attention mechanism (see
quantitative analysis on the Figure \ref{fig:actmaps} and Figure
\ref{fig:top10}).

\subsection{Performance evaluation on CPU}

Performance is a crucial point for practical re-identification models.  Modern
multi-camera multi-target person tracking (MCMT) approaches use
tracking-by-detection and hierarchical feature clustering paradigms that compute
re-identification features for each detected person in multiple video streams
\cite{zhang2017multitarget, Li2019StateawareRF}. Fast and accurate re-id model
in combination with real-time detector would allow to achieve close to real-time
performance in MCMT. In this work we provide a number of models demonstrating a
trade-off between cross-domain re-identification accuracy and performance on
CPU.  Table \ref{tab:perf} shows the performance numbers.  Any approach based on
ResNet-50 is slower than OSNet-IAP 1.0x by a factor of two. Our top-performing
model OSNet-IAP 1.0x is slightly slower than the original OSNet 1.0x because of
the adaptive pooling and Instance Normalization.  The inference of our lightest
model OSNet-IAP 0.25x takes only 1ms. That allows processing about 30 persons in
real time while outperforming much heavier approaches OSNetx1.0-IBN
\cite{zhou2019osnet} and MAR \cite{Yu2019UnsupervisedPR} on Market1501 (see
Table \ref{tab:sota}).

\section{Conclusion}

In this work we showcased the effectiveness of the metric-learning approach in
the cross domain person re-identification task. We proposed to use the
margin-based loss to extract discriminative features and showed that this
approach substantially improves cross-domain generalization. Also, we utilized a
number of training techniques and architecture adjustments to further boost our
results to the state-of-the-art level while maintaining compact size of our
model. Extensive experiments validated the effectiveness of the proposed method
as well as each training component individually.  We proved the real-time
performance of our approach and compared it against widespread ResNet-50-based
solution by benchmarking on CPU.

\bibliographystyle{IEEEtran}
\bibliography{IEEEabrv,egbib}
%
%
%

\end{document}